\documentclass[10pt,letterpaper]{article}
\usepackage[top=0.85in,footskip=0.5in,marginparwidth=2in]{geometry}

\usepackage[utf8]{inputenc}

\usepackage{cite}

\usepackage{nameref}

\usepackage[right]{lineno}

\usepackage{microtype}
\DisableLigatures[f]{encoding = *, family = * }

\raggedright
\usepackage{changepage}

\usepackage[aboveskip=1pt,labelfont=bf,labelsep=period,singlelinecheck=off]{caption}

\usepackage{mathtools}
\usepackage{bm}
\usepackage{amssymb}

\makeatletter
\renewcommand{\@biblabel}[1]{\quad#1.}
\makeatother

\usepackage{lastpage,fancyhdr,graphicx}
\usepackage{epstopdf}
\fancyheadoffset[L]{0.25in}
\fancyfootoffset[L]{0.25in}

\usepackage{color}

\definecolor{Gray}{gray}{.25}

\usepackage{graphicx}

\usepackage{ragged2e}
\usepackage{framed,multirow}

\usepackage{amssymb}
\usepackage{latexsym}
\usepackage{makecell}

\usepackage{url}
\usepackage{xcolor}

\usepackage[colorlinks=true]{hyperref}
\usepackage{algorithm}
\usepackage[noend]{algorithmic}
\usepackage{booktabs,subcaption,amsfonts,dcolumn}
\usepackage{float}
\usepackage{ulem}
\definecolor{newcolor}{rgb}{.8,.349,.1}
\definecolor{revision}{rgb}{0.0, 0.0, 0.0}

\begin{document}
\vspace*{0.35in}

\begin{flushleft}
{\Large
\textbf{Uncertainty-aware multi-view co-training for semi-supervised medical image segmentation and domain adaptation}\footnote{Preprint, to appear in Medical Image Analysis (\url{https://doi.org/10.1016/j.media.2020.101766})}
}
\newline
\\
Yingda Xia\textsuperscript{a},
Dong Yang\textsuperscript{b},
Zhiding Yu\textsuperscript{b},
Fengze Liu\textsuperscript{a},
Jinzheng Cai\textsuperscript{c},
Lequan Yu\textsuperscript{d},
Zhuotun Zhu\textsuperscript{a},
Daguang Xu\textsuperscript{b},
Alan Yuille\textsuperscript{a},
Holger Roth\textsuperscript{b,}\footnote{Corresponding author, email: hroth@nvidia.com},
\\
\bigskip
\textsuperscript{a} Johns Hopkins University
\\
\textsuperscript{b} Nvidia
\\
\textsuperscript{c} University of Florida
\\
\textsuperscript{d} The Chinese University of Hong Kong
\\
\end{flushleft}

\justify
\section*{Abstract}

Although having achieved great success in medical image segmentation, deep learning-based approaches usually require large amounts of well-annotated data, which can be extremely expensive in the field of medical image analysis. Unlabeled data, on the other hand, is much easier to acquire. Semi-supervised learning and unsupervised domain adaptation both take the advantage of unlabeled data, and they are closely related to each other. In this paper, we propose \textbf{uncertainty-aware multi-view co-training} (UMCT), a unified framework that addresses these two tasks for volumetric medical image segmentation. Our framework is capable of efficiently utilizing unlabeled data for better performance. We firstly rotate and permute the 3D volumes into multiple views and train a 3D deep network on each view. We then apply co-training by enforcing multi-view consistency on unlabeled data, where an uncertainty estimation of each view is utilized to achieve accurate labeling. Experiments on the NIH pancreas segmentation dataset and a multi-organ segmentation dataset show state-of-the-art performance of the proposed framework on semi-supervised medical image segmentation. Under unsupervised domain adaptation settings, we validate the effectiveness of this work by adapting our multi-organ segmentation model to two pathological organs from the Medical Segmentation Decathlon Datasets. Additionally, we show that our UMCT-DA model can even effectively handle the challenging situation where labeled source data is inaccessible, demonstrating strong potentials for real-world applications.

\begin{figure*}[!t]
\begin{center}
    \includegraphics[width=15.5cm]{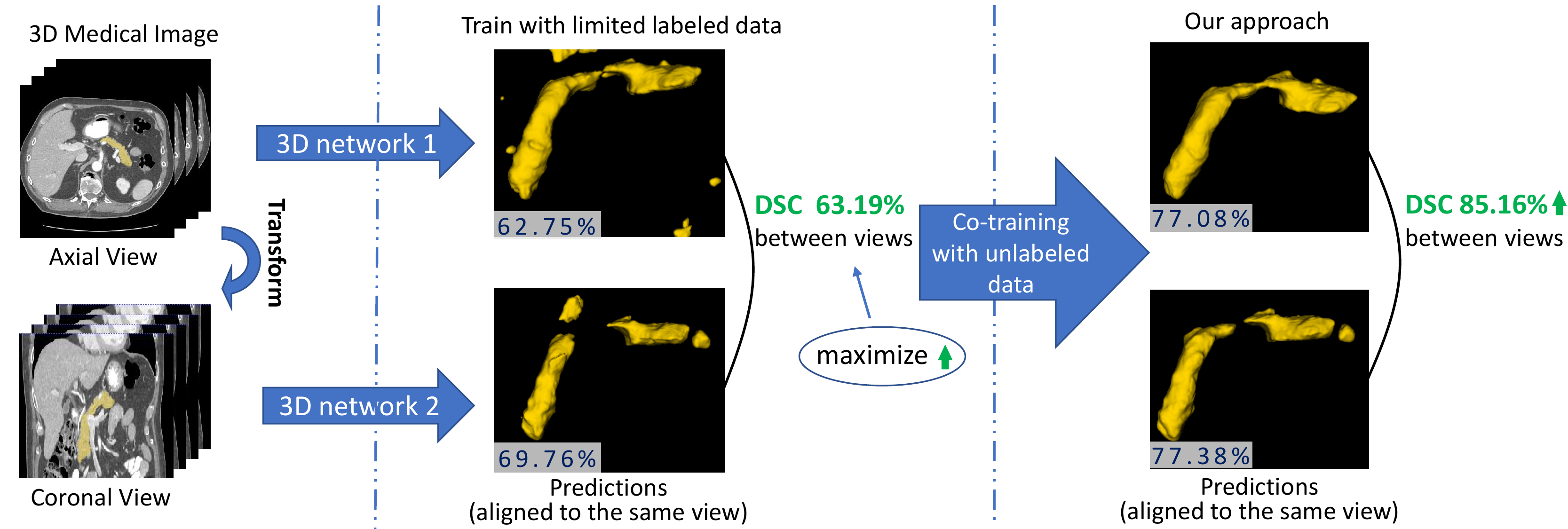}
\end{center}
\caption{
    An example of our approach for pancreas segmentation (best viewed in color). With limited training data, two 3D networks which are trained on axial and coronal view, respectively, both perform poorly as measured by DSC scores (in dark blue) with ground truth annotations. We observe that the DSC between the two views (in green) is also low, indicating large view differences. With our co-training approach, we minimize the difference between the two predictions on unlabeled data, resulting in significant improvement on each view.
}
\label{Fig:intro}
\end{figure*}

\section{Introduction}
Deep learning has achieved great successes in various computer vision tasks, such as 2D image recognition~\cite{krizhevsky2012imagenet,simonyan2014very,szegedy2015going,he2016deep,huang2017densely} and semantic segmentation~\cite{long2015fully,chen2018deeplab,zhao2017pyramid,chen2017rethinking}. However, deep networks usually rely on large-scale labeled datasets for training. When it comes to medical volumetric data, human labeling can be extremely costly and often requires expert domain knowledge. Medical image segmentation (i.e. the labeling tissues and organs in CTs and MRIs) plays a critical role in biomedical image analysis and surgical planning. Deep learning-based approaches have been widely adopted for this task and have led to state-of-the-art performance~\cite{ronneberger2015u, milletari2016v, yu2017volumetric, yu2017recurrent}. However, acquiring well-annotated segmentation labels in medical images requires both high-level expertise of radiologists and careful manual labeling of object masks or surface boundaries.

In this paper, we aim to design an approach that can utilize large-scale unlabeled data to improve volumetric medical image segmentation, and is applicable to the scenarios of both semi-supervised learning (SSL) and unsupervised domain adaptation (UDA). SSL and UDA share a common setting by assuming the availability of a labeled training set (denoted as $\mathcal{S}$), as well as an unlabeled one (denoted as $\mathcal{U}$). The difference between the two tasks is that for $\mathcal{S}$ and $\mathcal{U}$ we assume the same distribution in SSL while a larger domain shift is assumed in the UDA setting. Despite such differences, approaches in these two tasks are often closely related. SSL approaches such as self-training~\cite{selftraining1, selftraining2, bai2017semi}, co-training~\cite{blum1998combining, zhou2005semi} and GAN based methods~\cite{badgan,GANsemi} have been widely applied to UDA~\cite{cotraining-uda, cycada, adaptsegnet, dirt-t, self-ens, cbst, crst}, and vice versa.

Inspired by the success of co-training~\cite{blum1998combining} and its application to single 2D images~\cite{qiao2018deep}, we further extend this idea to 3D volumetric data. Typical co-training requires at least two views (i.e. sources) of the data, of which either should be sufficient to train a classifier on. Co-training minimizes the disagreement by assigning pseudo labels between each view on unlabeled data. \cite{blum1998combining} further proved that co-training has PAC-like guarantees on semi-supervised learning with an additional assumption that the two views are conditionally independent given the category. Since most computer vision tasks have only one source of data, encouraging view differences is a crucial factor for successful co-training. For example, \textit{deep co-training}~\cite{qiao2018deep} trains multiple deep networks to act as different views by utilizing adversarial examples~\cite{Goodfellow2015} to address this issue. Another aspect of co-training to emphasize is view confidence estimation. In multi-view settings, with growing differences between each view, the quality of each prediction becomes less and less guaranteed and might result in bad pseudo labels that can be harmful if used in the training process. Co-training could benefit from trusting reliable predictions and degrading the unreliable ones. However, distinguishing reliable and unreliable predictions is challenging for unlabeled data due to lack of ground-truth.

\begin{figure*}[!t]
\begin{center}
    \includegraphics[width=16cm]{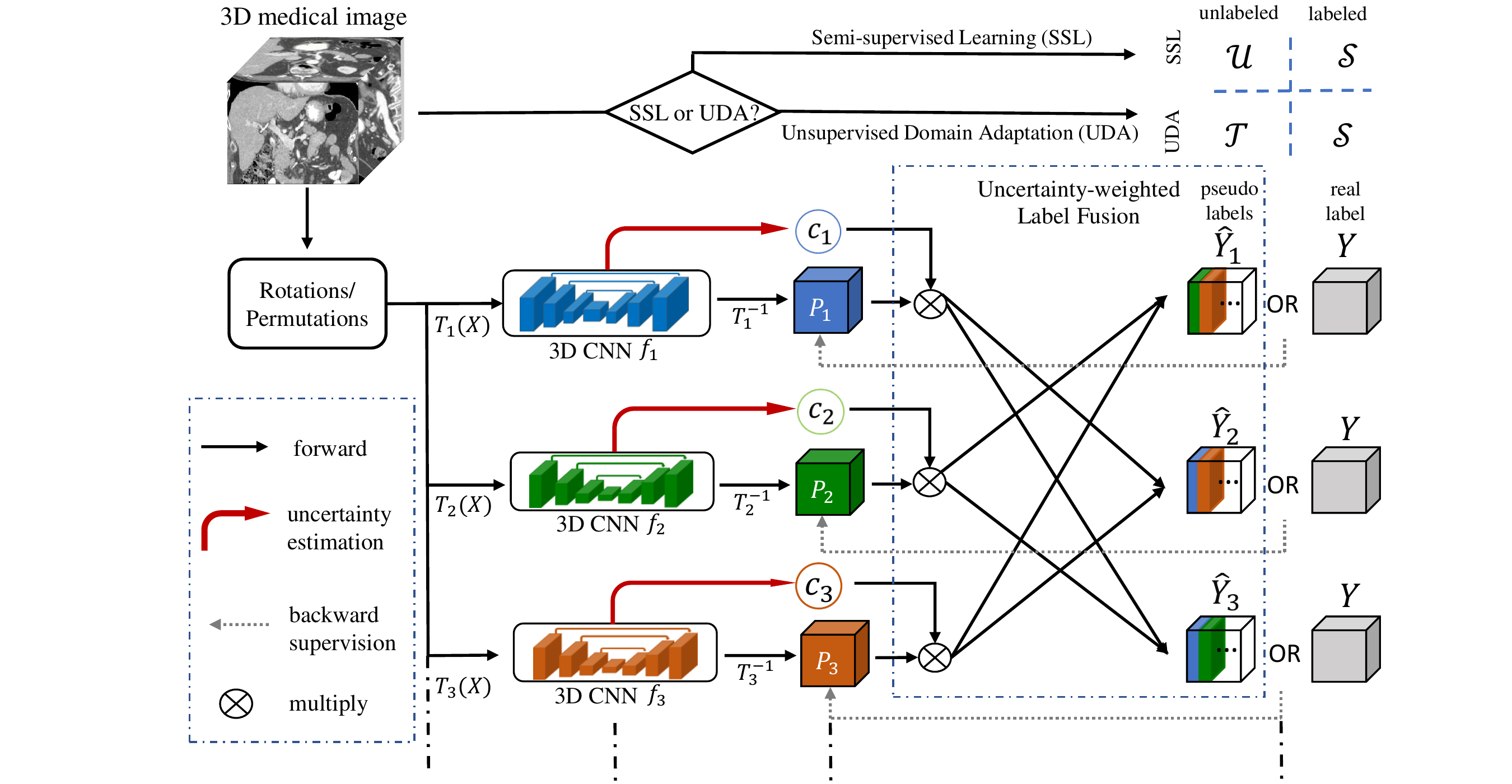}
\end{center}
\caption{
    Overall framework of \textbf{uncertainty-aware multi-view co-training} (UMCT), best viewed in color. \textcolor{revision}{UMCT can be applied to either the semi-supervised learning (SSL) task or the unsupervised domain adaptation (UDA) task, both of which include an unlabeled and a labeled subset of data. The overall pipeline is described as follows.} The $n$ multi-view inputs of $\mathbf{X}$ are first generated through different transforms $\mathbf{T}$, like rotations and permutations, before being fed into $n$ deep networks with asymmetrical 3D kernels. A confidence score $c$ is computed for each view by uncertainty estimation and acts as the weights to compute the pseudo labels $\hat{Y}$ of other views (Eq. \ref{Eqn:pseudo-label}) after inverse transform $\mathbf{T}^{-1}$ of the predictions. The pseudo labels $\hat{Y}$ for unlabeled data and ground truth $Y$ for labeled data are used as supervisions during training.
}
\label{Fig:Overall}
\end{figure*}

To address the above two important issues, we propose an \textit{uncertainty-aware multi-view co-training} (UMCT) framework, shown in Fig.~\ref{Fig:Overall}. 
We introduce view differences by exploring multiple viewpoints of 3D data through spatial transformations, such as rotation and permutation. \textcolor{revision}{The permutation here is defined as the rearrangements of the coordinate system, such as transpose and flip, and ``view" is defined as the transformed input data after permutation.} Hence, our multi-view approach naturally applies to analyzing 3D data and can be integrated with the proposed co-training framework. Fig.~\ref{Fig:intro} gives an example of the intuition of our approach in two-view scenario. On unlabeled data, we propose to maximize the similarity of the predictions between the two views, resulting in improved segmentation performance on each view.
Another key component is the view confidence estimation. We propose to estimate the uncertainty of predictions in each view with Bayesian deep networks by adding dropout in the architectures~\cite{gal2016dropout}. A confidence score is computed based on epistemic uncertainty~\cite{kendall2017uncertainties}, which can act as a weight for each prediction. After propagation through this \textit{uncertainty-weighted label fusion module} (ULF), a set of more accurate pseudo labels can be obtained for each view, which is used as supervision signal for unlabeled data. 

UMCT was previously published as a conference paper~\cite{UMCT}, in which we verified its effectiveness under standard semi-supervised settings on individual organs. In this paper, we extensively validate our approach on more challenging tasks, e.g. multi-organ segmentation. Moreover, we apply our approach to the task of unsupervised domain adaptation, with a labeled source domain and unlabeled target domain. In medical image analysis, this is considered as an important task since we should prefer a model or an approach that has the capability to generalize across datasets from different data sources (e.g. types of machines, acquisition protocols, and characteristics of patients). In addition to the original experiments on NIH pancreas dataset~\cite{roth2015deeporgan}, we validate our approach on a multi-organ dataset used in \cite{densevnet} with 8 labeled abdominal organs under semi-supervised settings. We then utilize our co-training approach to adapt the multi-organ model to the Medical Image Decathlon (MSD)~\cite{msd}) pathological liver and pancreas datasets. We even push our approach one step further by assuming that we only have the source model in the absence of source data. With very simple modifications, our final model UMCT-DA illustrates strong potential on this challenging scenario.

\section{Related Work}
\noindent
\textbf{Semi-supervised learning } aims at learning models with limited labeled data and a large proportion of unlabeled data~\cite{blum1998combining, zhou2005tri, belkin2006manifold, zhou2005semi}. Emerging semi-supervised approaches have been successfully applied to image recognition using deep neural networks~\cite{bachman2014learning,rasmus2015semi,sajjadi2016regularization,laine2017temporal,mean-teacher,miyato2018virtual,chen2018tri}. These algorithms mostly rely on additional regularization terms to train the networks to be resistant to some specific noise. A recent approach~\cite{qiao2018deep} extended the co-training strategy to 2D deep networks and multiple views, using adversarial examples to encourage view differences to boost performance. 

\noindent
\textbf{Semi-supervised medical image analysis. }
\cite{cheplygina2018not} mentioned that current semi-supervised medical analysis methods fall into 3 types - self-training (teacher-student models), co-training (with hand-crafted features) and graph-based approaches (mostly applications of graph-cut based optimization). \cite{bai2017semi} introduced a deep network based self-training framework with conditional random field (CRF) based iterative refinements for medical image segmentation. \cite{zhou2018semi} trained three 2D networks from three planar slices of the 3D data and fused them in each self-training iteration to get a stronger student model. \cite{li2018semi} extended the self-ensemble approach $\pi$ model~\cite{laine2017temporal} with 90-degree rotations making the network rotation-invariant. Generative adversarial network (GAN) based approaches are also popular recently for medical imaging~\cite{mic1,mic2,mic3}. \textcolor{revision}{Moreover, mixed supervisions~\cite{shah2018ms, mlynarski2019deep} combining dense label masks and weak labels like bounding boxes, slice- or image-level labels, etc., is another field of study to alleviate labeling efforts and is related to semi-supervised medical image analysis.} 

\noindent
\textbf{Uncertainty Estimation. }
Traditional approaches include particle filtering and CRFs~\cite{blake1993framework,he2004multiscale}. For deep learning, uncertainty is more often measured with Bayesian deep networks~\cite{gal2016dropout,gal2016uncertainty,kendall2017uncertainties}. In our work, we emphasize the importance of uncertainty estimation in semi-supervised learning, since most of the training data here is not annotated. We propose to estimate the confidence of each view in our co-training framework via Bayesian uncertainty estimation.

\noindent
\textbf{2D/3D hybrid networks. }
2D networks and 3D networks both have advantages and limitations. The former benefits from 2D pre-trained weights and well-studied architectures on natural images, while the latter better explores 3D information utilizing 3D convolution kernels. \cite{xia2018bridging,li2017h} either uses 2D probability maps or 2D feature maps for building 3D models. \cite{liu20173d} proposed a 3D architecture which can be initialized by 2D pre-trained models. Moreover, \cite{roth2018spatial,zhou2017fixed} illustrates the effectiveness of multi-view training on 2D slices, even by simply averaging multi-planar results, indicating complementary latent information exists in the biases of 2D networks. This inspired us to train 3D multi-view networks with 2D initializations jointly using an additional loss function for multi-view networks which encourages each network to learn from one another.

\noindent
\textbf{Unsupervised Domain Adaptation. } 
Contrary to semi-supervised learning, domain adaptation problems often contain two datasets that have different distribution. Under unsupervised domain adaptation (UDA) settings, networks are trained from a labeled source domain and an unlabeled target domain. Traditional approaches~\cite{uda1,uda2,uda3,uda4} align domains with statistical constraints. Recent works~\cite{uda-adv1,uda-adv2,cycada,adaptsegnet,fcninthewild,dirt-t,self-ens,cbst,crst} utilizes adversarial training and self-training to adapt feature training between source domain and target domain. In the field of medical image analysis, \cite{uda-med1,uda-med2,uda-med3} have investigated this topic with existing approaches, i.e. adversarial training and self-training.



\section{Problem Definitions}
\label{Sec:prob}
Before we describe our proposed approach, we firstly discuss the definition and relationships of the three problems, i.e. semi-supervised learning (SSL), unsupervised domain adaptation (UDA) and UDA without data from source domain. Table~\ref{Tab:task} lists the comparison among the three problems.

\noindent
\textbf{Semi-supervised learning.}
Under standard semi-supervised learning (SSL) settings, we denote $\mathcal{S}$ and $\mathcal{U}$ as the labeled and unlabeled dataset, respectively. Let $\mathcal{D} = \mathcal{S} \cup \mathcal{U}$ be the whole available dataset. We denote each labeled data pair as $(\mathbf{X}, \mathbf{Y}) \in \mathcal{S}$ and unlabeled data as $\mathbf{X} \in \mathcal{U}$. We aim to improve performance on a specific task with unlabeled data. When we consider volumetric medical image segmentation, $\mathbf{X}$ is a three-dimensional tensor and the ground truth $\mathbf{Y}$ is a densely-labeled voxel-wise 3D segmentation mask. 

\noindent
\textbf{Unsupervised domain adaptation} (UDA) assumes a labeled source domain dataset $\mathcal{S}$ and an unlabeled target domain dataset $\mathcal{T}$, where distributions of data are different but tasks are identical. Our goal is to achieve relatively high performance of a specific task on the target domain. In medical image analysis, domain gaps can result from differences in imaging modalities (e.g. CT / MRI / PET), qualities or imaging protocols (e.g. various machine types and doses of radiation), types of patients (e.g. healthy or with disease), and combinations thereof. The difference between UDA and SSL only lies in data distributions, so SSL approaches can also be applied to solve UDA problems. In our paper, we illustrate that our proposed approach can effectively handle both problems.

\noindent
\textbf{UDA without data from source domain} was barely investigated in the literature but is an important challenge to be addressed in the field of medical imaging. Here we assume an available pre-trained model from the source domain and unlabeled data from target domain. Differently from UDA, data from source domain is absent. 

In our work, we aim to propose a unified approach that is capable of solving the three tasks described above.

\begin{table}[htb]
\centering
\begin{tabular}{|l|c|c|c|c|}    
\hline
\thead{settings} & \thead{ dataset 1}&\thead{dataset 2} & \thead{same \\ domain?} & \thead{pre-trained \\ model?} \\
\hline\hline
SSL  &  labeled & unlabeled& yes & -\\
UDA & labeled & unlabeled& no & -\\
UDA w/o $\mathcal{S}$ &  N/A & unlabeled & no & on dataset 1\\
\hline
\end{tabular}
\caption{
    The relationship among the three settings i.e. semi-supervised learning (SSL), unsupervised domain adaptation (UDA) and UDA without source domain (UDA w/o $\mathcal{S}$).
}
\label{Tab:task}
\end{table}

\section{Uncertainty-aware Multi-view Co-training}
In this section, we introduce our framework of \textit{uncertainty-aware multi-view co-training} (UMCT) for semi-supervised segmentation and domain adaptation. UMCT is designed to effectively utilize unlabeled data, which is firstly targeted at semi-supervised segmentation of volumetric medical images. In the following sections, we will explain how they are achieved in our 3D framework: a general mathematical formulation of the approach is shown in Sec \ref{sec:overall}; then we demonstrate how to encourage view differences in Sec~\ref{sec:difference}, and how to compute the confidence of each view by uncertainty estimation in Sec \ref{sec:uncertainty}, which are the two factors to boost the performance of co-training. Last but not least, the UMCT-DA model is introduced in Sec \ref{sec:umct-da} for unsupervised domain adaptation.

\subsection{Overall Framework}
\label{sec:overall}
We first consider the task of semi-supervised segmentation for 3D data. Recall that $\mathcal{S}$ and $\mathcal{U}$ are the labeled and unlabeled set, respectively. Each labeled data pair is denoted as $(\mathbf{X}, \mathbf{Y}) \in \mathcal{S}$ and unlabeled data as $\mathbf{X} \in \mathcal{U}$. 
\textcolor{revision}{The ground truth $\mathbf{Y}$ is a voxel-wise segmentation label map which has the same shape as $\mathbf{X}$.}

Suppose for each input $\mathbf{X}$, we can generate $N$ different views of 3D data by applying a transformation $T_i$ (rotation or permutation), resulting in multi-view inputs $T_i(X)$, $i = 1,...,N$. Such operations will introduce a data-level view difference. $N$ models $f_i(\cdot)$, $i = 1,...,N$ are then trained over each view of data respectively. For $(\mathbf{X}, \mathbf{Y}) \in \mathcal{S}$, a supervised loss function $\mathcal{L}_{sup}$ is optimized to measure the similarity between the prediction of each view $p_i(\mathbf{X}) = T_i^{-1}\circ f_i\circ T_i(\mathbf{X})$ and $\mathbf{Y}$: 
\begin{equation}
\label{Loss1}
\mathcal{L}_{sup}(\mathbf{X}, \mathbf{Y}) = \sum_{i=1}^N \mathcal{L}(p_i(\mathbf{X}), \mathbf{Y}),
\end{equation}

\textcolor{revision}{where $\mathcal{L}$ is a standard loss function for segmentation tasks and $\{p_i(\mathbf{X})\}_{i=1}^N$ are the corresponding voxel-wise prediction score maps after inverse rotation or permutation.}


For unlabeled data, we make a co-training assumption under a semi-supervised setting. The co-training strategy assumes the predictions on each view should reach a consensus. So the prediction of each model can act as a pseudo label to supervise other views in order to learn from unlabeled data. However, since the prediction of each view is expected to be diverse after encouraging the view differences, the quality of each view's prediction needs to be measured before generating trustworthy pseudo labels. This is accomplished via \textit{uncertainty-weighted label fusion module} (ULF) introduced in Sec~\ref{sec:uncertainty}. With ULF, the co-training loss for unlabeled data can be formulated as:
\begin{equation}
\label{Loss2}
\mathcal{L}_{cot}(\mathbf{X}\textcolor{revision}{, \hat{\mathbf{Y_i}}}) = \sum_i^N \mathcal{L}(p_i(\mathbf{X}), \hat{\mathbf{Y_i}}), 
\end{equation}
where
\begin{equation}
    \hat{\mathbf{Y_i}} = U_{f_1,..f_n}(p_1(\mathbf{X}),..,p_{i-1}(\mathbf{X}),p_{i+1}(\mathbf{X}),..,p_n(\mathbf{X}))
\label{eqn:pl}
\end{equation}
 is the pseudo label for the $i^{th}$ view, $U_{f_1,..f_n}$ is the ULF computational function, which we  will further explain in Sec~\ref{sec:uncertainty}.

Overall, the combined loss function is:
\begin{equation}
\label{loss3}
\sum_{(\mathbf{X},\mathbf{Y}) \in \mathcal{S}} \mathcal{L}_{sup}(\mathbf{X}, \mathbf{Y}) + \lambda_{cot} \sum_{\mathbf{X}\in\mathcal{U}}\mathcal{L}_{cot}(\mathbf{X}\textcolor{revision}{, \hat{\mathbf{Y_i}}}).
\end{equation}
where $\lambda_{cot}$ is a tunable weight coefficient.

\begin{algorithm}[t]
\caption{Uncertainty-aware Multi-view Co-training}
\textbf{Input:}\\ 
Labeled dataset $\mathcal{S}$ \& Unlabeled dataset $\mathcal{U}$\\
\textit{uncertainty-weighted label fusion module} (ULF) $U_{f_1,..f_n}(\cdot)$\\
\textbf{Output:}\\
Model of each view $f_1,..f_n$
\begin{algorithmic}[1]
\WHILE{stopping criterion not met}
\STATE Sample batch $b_l = (x_l, y_l) \in \mathcal{S}$ and batch $b_u=(x_u) \in \mathcal{U}$
\STATE Generate multi-view inputs $T_i(x_l)$ and $T_i(x_u)$, $i \in \{1, .., N\}$
\FOR{i \textbf{in} all views}
\STATE Compute predictions for each view and apply inverse rotation or permutation\\ $p_i(x_l) \gets T_i^{-1}\circ f_i\circ T_i(x_l)$\\$p_i(x_u) \gets T_i^{-1}\circ f_i\circ T_i(x_u)$
\ENDFOR
\FOR{i \textbf{in} all views}
\STATE Compute pseudo labels for $x_u$ with ULF \\ $\hat{y_i} \gets U_{f_1,..f_n}(p_1(x_u),..,p_{i-1}(x_u), p_{i+1}(x_u),..,p_n(x_u)) $
\ENDFOR
\STATE $\mathcal{L}_{sup} = \frac{1}{|b_l|}\sum_{(x_l,y_l)\in b_l}[\sum_i^N\mathcal{L}(p_i(x_l), y_l)]$
\STATE $\mathcal{L}_{cot} = \frac{1}{|b_u|}\sum_{(x_u)\in b_u}[\sum_i^N\mathcal{L}(p_i(x_u), \hat{y_i})]$
\STATE $\mathcal{L} = \mathcal{L}_{sup} + \lambda_{cot}\mathcal{L}_{cot}$
\STATE Compute gradient of loss function $\mathcal{L}$ and update network parameters $\{\theta_i\}$ by back propagation
\ENDWHILE
\RETURN $f_1,..f_n$
\end{algorithmic}
\label{algo1}
\end{algorithm}

\subsection{Encouraging View Differences}
\label{sec:difference}
A successful co-training requires the ``views" to be different in order to learn complementary information in the training procedure. 
In our framework, several techniques are proposed to encourage view differences, both at the data level and the feature level of the neural networks.

\noindent
\textbf{3D multi-view generation. } As stated above, in order to generate multi-view data, we transpose $\mathbf{X}$ into multiple views by rotations or permutations\footnote{A permutation rearranges the dimensions of an array in a specific order.} $\mathbf{T}$. For three-view co-training, these can correspond to the coronal, sagittal and axial views in medical imaging, which matches the multi-planar reformatted views that radiologists typically use to analyze the image. Such operation is a natural way to introduce data-level view difference.

\noindent
\textbf{Asymmetric 3D kernels and 2D initialization. }
The co-training assumption encourages models to make similar predictions on both $\mathcal{S}$ and $\mathcal{U}$, which potentially can lead to collapsed neural networks mentioned in~\cite{qiao2018deep}\textcolor{revision}{, a phenomenon that results in a sudden and significant drop in validation accuracy during training of co-training algorithms. In our multi-view settings, this could also happen when the models from different views only learn the permutation or rotation of the kernels, resulting in exactly the same learned feature representation despite the view-point difference.} To address this problem, we further encourage view difference at the feature level by designing a task-specific model. We propose to use asymmetric 3D models initialized with 2D pre-trained weights as the backbone network of each view to encourage diverse features for each view learning. 
\textcolor{revision}{In practice, we modify the symmetric 3D convolutional kernels $n\times n \times n$ into $n\times n\times 1$ for each branch after the permutation to avoid learning symmetrical representations among views.}
This structure also makes the model convenient to be initialized with 2D pre-trained weights but fine-tuned in a 3D fashion. 

\subsection{Compute Reliable Psuedo Labels for Unlabeled Data with Uncertainty Estimation}
\label{sec:uncertainty}
Encouraging view difference means enlarging the variance of each view's prediction $var(p_i(\mathbf{X}))$. This raises the question of which view we should trust most on unlabeled data during co-training. Bad predictions from one view may hurt the training procedure of other views through pseudo-label assignments. Meanwhile, encouraging to trust a good prediction as a ``strong'' label from co-training will boost the performance, and lead to improved performance of overall semi-supervised learning. Instead of assigning a pseudo-label for each view directly from the predictions of other views, we propose an adaptive approach, namely \textit{uncertainty-weighted label fusion module} (ULF), to fuse the outputs of different views. ULF is built up of all the views, takes the predictions of each view as input, and then outputs a set of pseudo labels for each view.

Motivated by uncertainty measurements in Bayesian deep networks, we measure the uncertainty of each view branch for each training sample after turning our model into a Bayesian deep network by adding dropout layers. Between the two types of uncertainty candidates -- aleatoric and epistemic uncertainties, we choose to compute the epistemic uncertainty that is driven by the lack of training data~\cite{kendall2017uncertainties}. Such measurement fits the semi-supervised learning goal: to improve the model's generalizability by exploring unlabeled data. Suppose $y$ is the output of a Bayesian deep network, then the epistemic uncertainty can be estimated as:

\begin{equation}
    U_e(y) \approx \frac{1}{K} \sum_{k = 1}^K \hat{y_k}^2 - (\frac{1}{K}\sum_{k=1}^K\hat{y_k})^2,
\end{equation}
where $\{ \hat{y}_k\}_{k=1}^K$ are a set of sampled outputs. \textcolor{revision}{These sampled outputs are obtained by feeding the same input volume into the sub-network defined by $K$ different random dropout configurations~\cite{kendall2017uncertainties}. The voxel-wise epistemic uncertainty is estimated as the statistical variance of the $K$ predictions. More details are available in Sec~\ref{impledetails}.}

With a transformation function $\mathbf{h(\cdot)}$, we can transform the uncertainty score into a confidence score $\mathbf{c}(y) = \mathbf{h}(U_e(y))$. In practice, we simply define $\mathbf{h}(U_e(y)) = 1 / U_e(y)$. After normalization over all views, the confidence score will act as the weight for each prediction to assign as a pseudo label for other views. The pseudo label $\hat{\mathbf{Y_i}}$ assigned for a single view $i$ can be formulated as 
\begin{equation}
	\hat{\mathbf{Y_i}} = \frac{\sum_{j\neq i}^N\mathbf{c}(p_j(\mathbf{X}))p_j(\mathbf{X})}{\sum_{j\neq i}^N\mathbf{c}(p_j(\mathbf{X}))}.
\label{Eqn:pseudo-label}
\end{equation}

\textcolor{revision}{Thus the pseudo label $\hat{\mathbf{Y_i}}$ for view $i$ is computed from predictions from all the other views.}

\subsection{UMCT-DA model for unsupervised domain adaptation}
\label{sec:umct-da}

\noindent
\textbf{Standard unsupervised domain adaptation (UDA)}

We extensively validate our approach on unsupervised domain adaptation setting, where the labeled source domain $\mathcal{S}$ and the unlabeled target domain $\mathcal{T}$ are available for training. The task is shared between the two domains and the ultimate goal is to achieve good performance on the target domain test data. Despite the domain shift in labeled and unlabeled data, the overall settings of semi-supervised learning (SSL) and unsupervised domain adaptation (UDA) are the same. Hence, we can directly apply our UMCT to solve this problem. The optimization objective can be modified as the follows :

\begin{equation}
\label{loss_uda}
\mathcal{L}_{UDA} = \sum_{(\mathbf{X},\mathbf{Y}) \in \mathcal{S}} \mathcal{L}_{sup}(\mathbf{X}, \mathbf{Y}) + \lambda_{cot} \sum_{\mathbf{X}\in\mathcal{T}}\mathcal{L}_{cot}(\mathbf{X}).
\end{equation}

where $\mathcal{S}$ is the labeled source domain and $\mathcal{T}$ is the unlabeled target domain. 

\vspace{0.3cm}
\noindent
\textbf{UDA without source domain data}

Standard UDA methods usually require the existence of source domain data to allow joint training while doing adaptation to the target domain. Here we consider a more challenging setting where source domain data is unavailable and only deep network model (denoted as $\mathbf{M_{\mathcal{S}}}$) pre-trained on source domain is available. In our co-training framework, when source data is unavailable, we can still finetune $\mathbf{M_{\mathcal{S}}}$ with $\mathcal{L}_{cot}$ by iteratively refining pseudo labels. The objective function for UDA without source domain data, namely \textbf{UMCT-DA}, can be formulated as:
\begin{equation}
\label{loss_uda_ns}
\mathcal{L}_{UDA} = \sum_{\mathbf{X} \in \mathcal{T}} \lambda_{cot} \mathcal{L}_{cot}(\mathbf{X}).
\end{equation}

\subsection{Implementation Details}
\label{impledetails}
\noindent
\textbf{Network Structure. }
In practice, we build an encoder-decoder network based on ResNet-18~\cite{he2016deep}, and modify it into a 3D version. For the encoder part, the first $7\times 7$ convolution layer is extended to $7\times 7 \times 3$ kernels for low-level 3D feature extraction similar to \cite{liu20173d}. All other $3\times 3$ convolution layers are simply changed into $3\times 3\times 1$ that can be trained as a 3D convolution layer. In the decoder part, we adopt 3 skip connections from the encoder followed by 3D convolutions to give low-level cues for more accurate boundary prediction needed in segmentation tasks.

\noindent
\textbf{Uncertainty-weighted Label Fusion. }
In terms of view confidence estimation, we modify the network into a Bayesian deep network by adding dropouts. We sample $K = 10$ outputs for each view and compute voxel-wise epistemic uncertainty. \textcolor{revision}{Since we are using Dice loss~\cite{milletari2016v}, a common loss function for medical image segmentation which is computed on the image level, an image-wise uncertainty estimation is most suitable. We thus sum over the whole volume to estimate the uncertainty for each view.} We then simply use the reciprocal for the confidence transformation function $\mathbf{h} \textcolor{revision}{= \frac{1}{\mathbf{c}}}$ to compute the confidence score. The resulting pseudo label assigned for each view is a weighted average of all predictions of multiple views based on the normalized confidence score.

\noindent
\textbf{Loss Function. }
We extend the Dice loss~\cite{milletari2016v} for multi-class targets as our training objective function: 
\begin{equation}
\textcolor{revision}{\mathcal{L}_{Dice} = \frac{1}{D}\sum_{d=0}^D (1 - \frac{2\sum_{i=1}^{N}y_i^d\hat{y}_i^d}{\sum_{i=1}^{N} (y_i^d)^2 + \sum_{i=1}^{N} (\hat{y}_i^d)^2}),}
\end{equation}

\noindent
\textbf{Data Pre-Processing. }
All the training and testing data are firstly re-sampled to an isotropic volume resolution of 1.0 $mm$ for each axis. Data intensities are normalized to have zero mean and unit variance. We adopt patch-based training, and sample training patches of size $96^3$ with $1$:$1$ ratio between foreground and background.

\noindent
\textbf{Training. }
Our training algorithm is shown in Algorithm~\ref{algo1}. 
We firstly train the views separately on the labeled data and then conduct our co-training by fine-tuning the weights. The stochastic gradient descent (SGD) optimizer is used in both stages. In the view-wise training stage, a constant learning rate policy at $7\times 10^{-3}$, momentum at $0.9$ and weight decay of $4\times 10^{-5}$ for 20k iterations is used. In the co-training stage, we adopt a constant learning rate policy at $1\times 10^{-3}$ and train for 5k iterations. \textcolor{revision}{The parameter $\lambda_{cot} = 0.2$ resulted in the best performance which we report here.} \textcolor{revision}{The batch size is 20 in co-training, among which 4 images are labeled and 16 are unlabeled, maintaining a ratio of labeled and unlabeled to be 1:4.} Our framework is implemented in PyTorch. \textcolor{revision}{For 3D ResNet-18 on NIH dataset, the whole co-training procedure takes $\sim$24 hours on one single NVIDIA Titan RTX GPU with 24 GB memory. In our implementation, training occupies $\sim$15GB GPU memory in total.}

\noindent
\textbf{Testing. }
In the testing phase, there are two choices to finalize the output results: either to choose one single view prediction or to ensemble the predictions of the multi-view outputs with majority voting. We will report both results in subsequent sections for fair comparisons with the baselines since the multiple view networks can be thought of being similar to the ensemble of several single view models. The experimental results show that our model improves the performance in both settings (single view and multi-view ensemble) \textcolor{revision}{over all the other approaches}. We use sliding-window testing and re-sample our testing results back to the original image resolution to obtain the final results. Testing time for each case ranges from $1$ minute to $5$ minutes depending on the size of the input volume.

\begin{table*}[!btp]
\centering
\begin{tabular}{|l|c|c|c|c|}    
\hline
Method & Backbone  & 10\% lab& 20\% lab\\
\hline\hline
Supervised & 3D ResNet-18  & 66.75 & 75.79\\
\hline\hline
DMPCT~\cite{zhou2018semi}  &  2D ResNet-101 & 63.45& 66.75\\
DCT~\cite{qiao2018deep} (2v) & 3D ResNet-18 & 71.43& 77.54\\
TCSE~\cite{li2018semi} &  3D ResNet-18 & 73.87 & 76.46\\
\hline\hline
Ours (2 views) &  3D ResNet-18 &75.63 & 79.77\\
Ours (3 views) &  3D ResNet-18 &77.55 & 80.14\\
Ours (6 views) &  3D ResNet-18 &77.87 & 80.35\\
\hline
\hline
Ours (ensemble) &  3D ResNet-18 &\textbf{78.77} & \textbf{81.18}\\
\hline
\end{tabular}
\caption{
    Comparison to other semi-supervised approaches on NIH dataset (DSC, \%).  Note that we use the same backbone network as~\cite{li2018semi}~\cite{qiao2018deep}. Here, ``2v" means two views. For our approach, we report the average of all single views' DSC score for a fair comparison (2 views to 6 views), as well as multi-view ensemble results. ``10\% lab"  and ``20\% lab" mean the percentage of labeled data used for training.
}
\label{Tab:semiTE}
\end{table*}

\section{Experiments}
 In this section, we first evaluate our framework under semi-supervised settings on the NIH pancreas segmentation dataset~\cite{roth2015deeporgan} with cases from a healthy patient population (e.g. kidney donors\footnote{\url{https://wiki.cancerimagingarchive.net/display/Public/Pancreas-CT}}); and an multi-organ segmentation dataset~\cite{densevnet} with eight abdominal organs~\cite{btcv} with conditions mostly unrelated to the organs of interest (e.g. colorectal cancer or ventral hernia\footnote{\url{https://www.synapse.org/\#!Synapse:syn3193805/wiki/217789}}). We will provide detailed experiments, including ablation studies, on the former dataset. \textcolor{revision}{Note that the volumes come from different patients in each dataset and were separated at the patient-level for the different training, validation and testing splits.} Next, we validate the capability of our approach on the task of unsupervised domain adaptation, which is critical but under-investigated in the field of medical image analysis. The multi-organ segmentation dataset serves as source data. The targets of adaptation include two pathological organ datasets i.e. pancreas and liver datasets in the Medical Segmentation Decathlon (MSD)~\cite{msd}, which both can include tumors in their respective organs. More strictly, we also evaluate our approach under the situation where source data is inaccessible (UDA without source data).

\begin{figure*}[h]
\begin{center}
    \includegraphics[width=15cm]{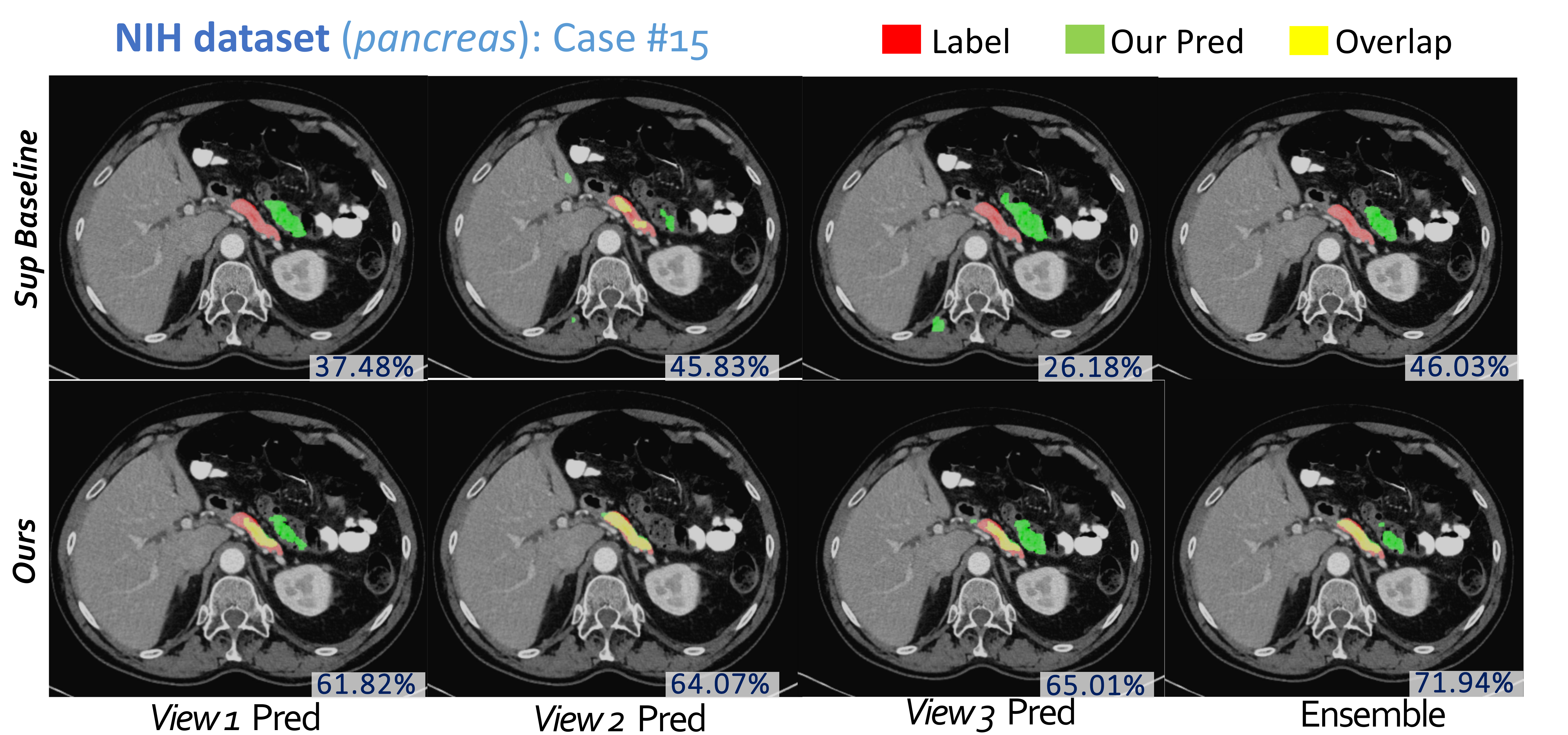}
\end{center}
\caption{
    2D visualizations for one example of NIH pancreas segmentation dataset 10\% labeled data setting. \textcolor{revision}{The first row is the supervised baseline and the second row is the prediction after our 3-view co-training.} DSC scores are largely improved. Best viewed in color.
}
\label{Fig:semi}
\end{figure*}

\subsection{NIH Pancreas Segmentation Dataset}

The NIH pancreas segmentation dataset contains 82 abdominal CT volumes. The width and height of each volume are 512, while the axial view slice number can vary from 181 to 466. Under semi-supervised settings, the dataset is randomly split into 20 testing cases and 62 training cases. We report the results of 10\% labeled training cases (6 labeled and 56 unlabeled), 20\% labeled training cases (12 labeled and 50 unlabeled) and 100\% labeled training cases. 

\subsubsection{Results}

\textcolor{revision}{In Table~\ref{Tab:semiTE}, we first report the average of all single views’ DSC score for a fair comparison (2 views to 6 views, last 2-4 rows), which can be viewed as the average performance of one single view model. Then we report the multi-view ensemble results (6 view ensemble, last row), where we align the multi-view prediction maps to the same view (axial) and average the prediction maps at each pixel to make a final prediction. For 2-view co-training, we use the axial and coronal views. For 3-view co-training, we use the axial, coronal and sagittal view. For 6-view co-training, we use the axial, coronal and sagittal view as well as the horizontal flip version of the three views ($3\times2=6$). The first row is the supervised training results, using only labeled data and trained on the axial view.}
The segmentation accuracy is evaluated by Dice-S{\o}rensen coefficient (DSC). A large margin improvement over the fully supervised baselines in terms of single view performance can be observed, proving that our approach effectively leverages the unlabeled data. A Wilcoxon signed-rank test comparing to the supervised baseline's results (20\% labeling) shows significant improvements of our approach with a $p$-value of 0.0022. Fig.~\ref{Fig:semi} shows 3 cases in 2D and 3D with ITK-SNAP~\cite{py06nimg}. In addition, our model is compared with the state-of-the-art semi-supervised approach of deep co-training~\cite{qiao2018deep} and recent semi-supervised medical segmentation approaches. In particular, we compare to \cite{li2018semi} who extended the $\pi$ model~\cite{laine2017temporal} with transformation consistent constraints; and \cite{zhou2018semi} who extended the self-training procedure by iteratively updating pseudo labels on unlabeled data using a fusion of three 2D networks trained on cross-sectional views. The results reported in Table~\ref{Tab:semiTE} are based on our careful re-implementations in order to allow a fair comparison.


The implementations of \cite{qiao2018deep} and \cite{li2018semi} are operated on the axial view of our single view branch with the same backbone structure (our customized 3D ResNet-18 model). Our co-training approach achieve about 4\% gain in the 10\% labeled and 90\% unlabeled settings. We also find that improvements of other approaches are small in the 20\% settings (only 1\% compared to the baseline), while ours still is capable to achieve a reasonable performance gain with the growing number of labeled data. For \cite{zhou2018semi} with a 2D approach, their experiment is conducted on 50 labeled cases. We modify their backbone network (FCN~\cite{long2015fully}) into DeepLab v2~\cite{chen2018deeplab}, in order to fit our stricter settings (6 and 12 labeled cases). This modification leads to an improvement of 3\% in 100\% fully supervised training (from 73\% to 76\%). Their approach outputs the result after using an ensemble \textcolor{revision}{with majority voting of three slice-wise 2D models obtained from their semi-supervised training approach.}. 

Since the main difference in two-view learning between our approach and \cite{qiao2018deep} is the way of encouraging view differences, the results illustrate the effectiveness of our multi-view analysis combined with asymmetric feature learning on 3D co-training. With more views, our uncertainty-weighted label fusion can further improve co-training performance. We will report ablation studies later in this section.

\subsubsection{Analysis and ablation studies}

\noindent
\textbf{Data utilization efficiency}

\begin{figure}[h]
\begin{center}
    \includegraphics[width=9cm]{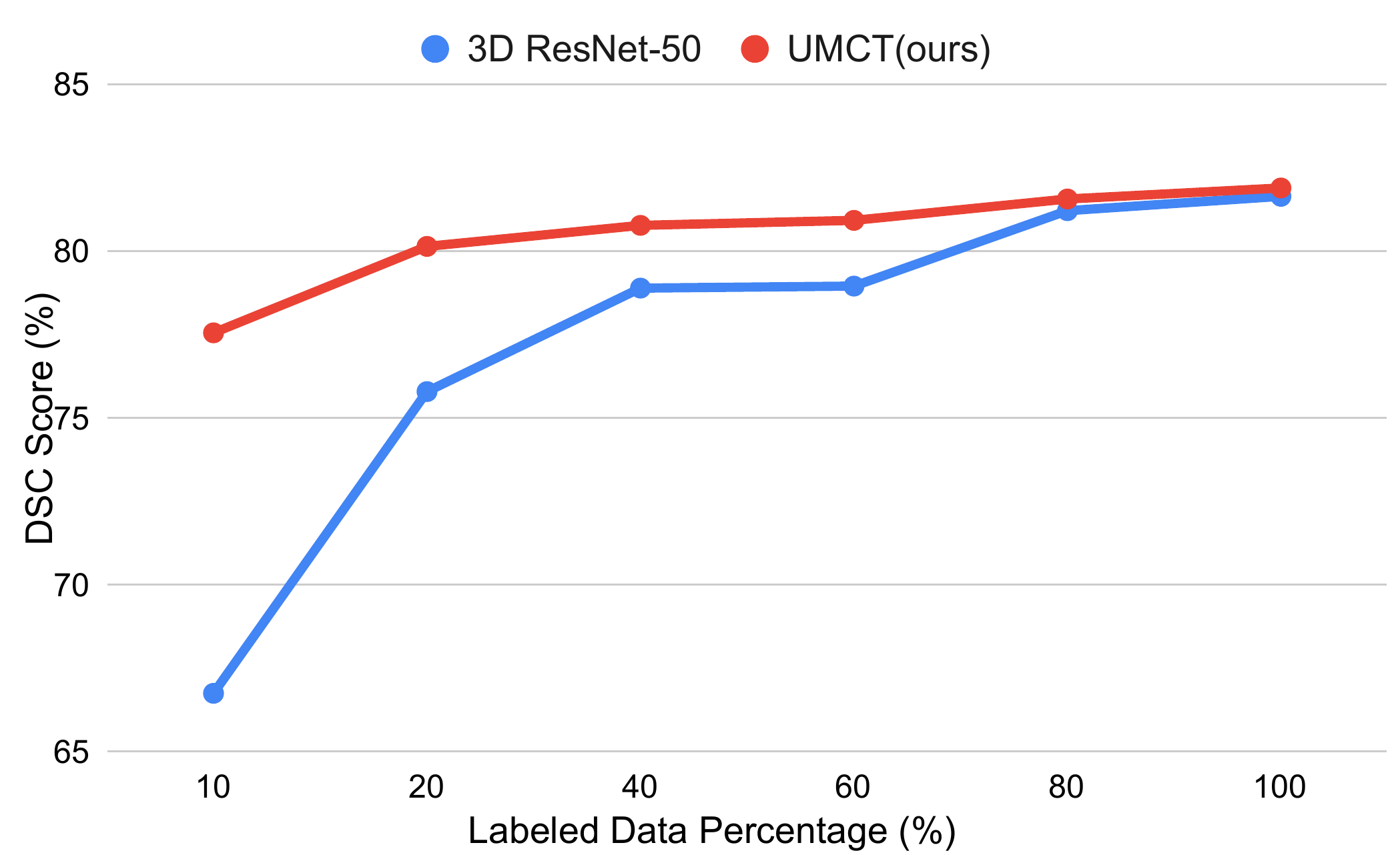}
\end{center}
\caption{
    Performance plot of our semi-supervised approach over the fully-supervised baseline on different labeled data ratio.
}
\label{Fig:percentage}
\end{figure}

We perform a study on data utilization efficiency of our approach compared to the baseline fully-supervised network (3D ResNet-18). Fig.~\ref{Fig:percentage} shows the performance change according to labeled data proportion on NIH pancreas segmentation. From the plot, one can see that when labeled data is over 80\%, simple supervised training (with 3D ResNet-18) suffices. Note that our approach with 20\% labeled data (DSC 80.35\%) performs better than 60\% supervised training (DSC 78.95\%). At such a performance, our approach can save $\sim$ 70\% of the labeling efforts.

\noindent\textbf{Effect of backbone structure}

Our backbone selection (2D-initialized, heavily asymmetric 3D architecture) will introduce 2D biases in the training phase while benefiting from such 2D pre-trained models. We have claimed that we can utilize the complementary information from 3-view networks while exploring the unlabeled data with UMCT. We give an ablation study on the network structure, which contains a V-Net~\cite{milletari2016v}, a common 3D segmentation network with all symmetrical kernels in all dimensions. Such network also shares a similar amount of parameters with our customized 3D ResNet-18, see Table~\ref{tab:ab_backbone}. The results of V-Net show that our multi-view co-training can be generally and successfully applied to 3D networks. Although the results of fully supervised parts are similar, our ResNet-18 outperforms V-Net by more than $1\%$, illustrating that our asymmetric design, encouraging view differences, brings advantages over traditional 3D deep networks.

\begin{table}[htp]
\centering
\begin{tabular}{|l|c|c|c|c|}    
\hline
Backbone & Params & MACs & 10\% Sup & Ours\\
\hline\hline
V-Net & 9.44M& 41.40G& 66.97 & 76.89	\\
3D ResNet-18 &11.79M &17.08G&66.76& 77.55\\
3D ResNet-50 & 27.09M& 23.03G&67.96& \textbf{78.74}\\
\hline
\end{tabular}
\caption{ Ablation studies on backbone structures (3 views UMCT). "Params" is short for parameters and "MACs" is short for multiply–accumulate operations. "10\% Sup" means supervised training with 10\% labeled data. \textcolor{revision}{A Wilcoxon signed-rank test reveals significant improvements ($p << 0.01$) of our 3D ResNets  over V-Net in the last column, illustrating our asymmetrical design is beneficial for our co-training method.}}
\label{tab:ab_backbone}
\end{table}

\noindent
\textbf{Uncertainty-weighted label fusion (ULF)}

ULF acts as an important role in pruning out bad predictions and keeping good ones as supervision to train other views. Table~\ref{tab:ab_ULF} gives the single view results in multiple views experiments.  The performance becomes better with more views. For two views, ULF is not applicable since we can only obtain one view prediction as a pseudo label for the other view. For three views and six views, ULF helps boost the performance, illustrating the effectiveness of our proposed approach for view confidence estimation. 

\begin{table}[htp]
\centering
\begin{tabular}{|l|c|}    
\hline
Views & DSC(\%)\\
\hline\hline
2 views & 75.63\\
\hline
3 views & 76.49\\
3 views + ULF & \textbf{77.55}\\
\hline
6 views & 76.94\\
6 views + ULF& \textbf{77.87}\\
\hline
\end{tabular}
\caption{
    On uncertainty-weighted label fusion (ULF) with difference views in training (10\% labeled data, 3D ResNet-18).
}
\label{tab:ab_ULF}
\end{table}

\begin{table*}[h]
\centering
\resizebox{\textwidth}{!}{%
\begin{tabular}{|l|>{\color{revision}}c|>{\color{revision}}c|>{\color{revision}}c|>{\color{revision}}c|>{\color{revision}}c|>{\color{revision}}c|>{\color{revision}}c|>{\color{revision}}c|}    
\hline
Experiment setups & \textcolor{black}{spleen}&	\textcolor{black}{l.kidney}&	\textcolor{black}{gallbladder}&	\textcolor{black}{esophagus}&	\textcolor{black}{liver}&	\textcolor{black}{stomach}	&\textcolor{black}{pancreas}&	\textcolor{black}{duodenum}\\
\hline
Supervised (upper bound)& 94.20&	93.90&	71.89&	66.74&	94.78&	88.60&	81.46&	71.29\\
\hline\hline
10\% lab & 88.46&	90.88&	42.77&	52.41&	91.33&	76.50&	69.63&	52.52 \\
10\% lab+90\% unlab (ours)  & \textbf{91.14}& \textbf{92.35}& \textbf{58.29} &	\textbf{57.61} &	\textbf{92.23}&	\textbf{79.67}&	\textbf{73.86}&	\textbf{57.50}\\
\hline\hline
20\% lab & 92.77&	92.29&	63.60&	61.84&	93.95&	82.56&	75.60&	60.26\\
20\% lab+80\% unlab (ours)  &  92.80&	\textbf{92.99}&	\textbf{66.29}&	\textbf{65.01}&	93.93&	\textbf{83.67}&	\textbf{77.91}&	\textbf{63.34}\\

\hline
\end{tabular}}
\caption{
    Experimental results for semi-supervised learning on a multi-organ dataset \textcolor{revision}{under four fold cross-validation}. "lab" is short for "labeled" and "unlab" is short for "unlabeled". \textcolor{revision}{Supervised results (first row) uses 100\% labeled training data in the training set, which is the upper bound but requires 100\% annotation. 10\% lab means we only use 10\% training data with annotation for supervised training. 10\%lab + 90\% unlab (ours) means we use 10\% labeled data and 90\% unlabeled data for our co-training method.} \textcolor{revision}{Results are reported via 4-fold cross-validation. Numbers in \textbf{bold} indicate significant improvement over supervised counterparts by Wilcoxon signed rank tests ($p<<0.01$).}
}
\label{tab:multiorgan}
\end{table*}

\begin{figure}[h]
\begin{center}
    \includegraphics[width=12cm]{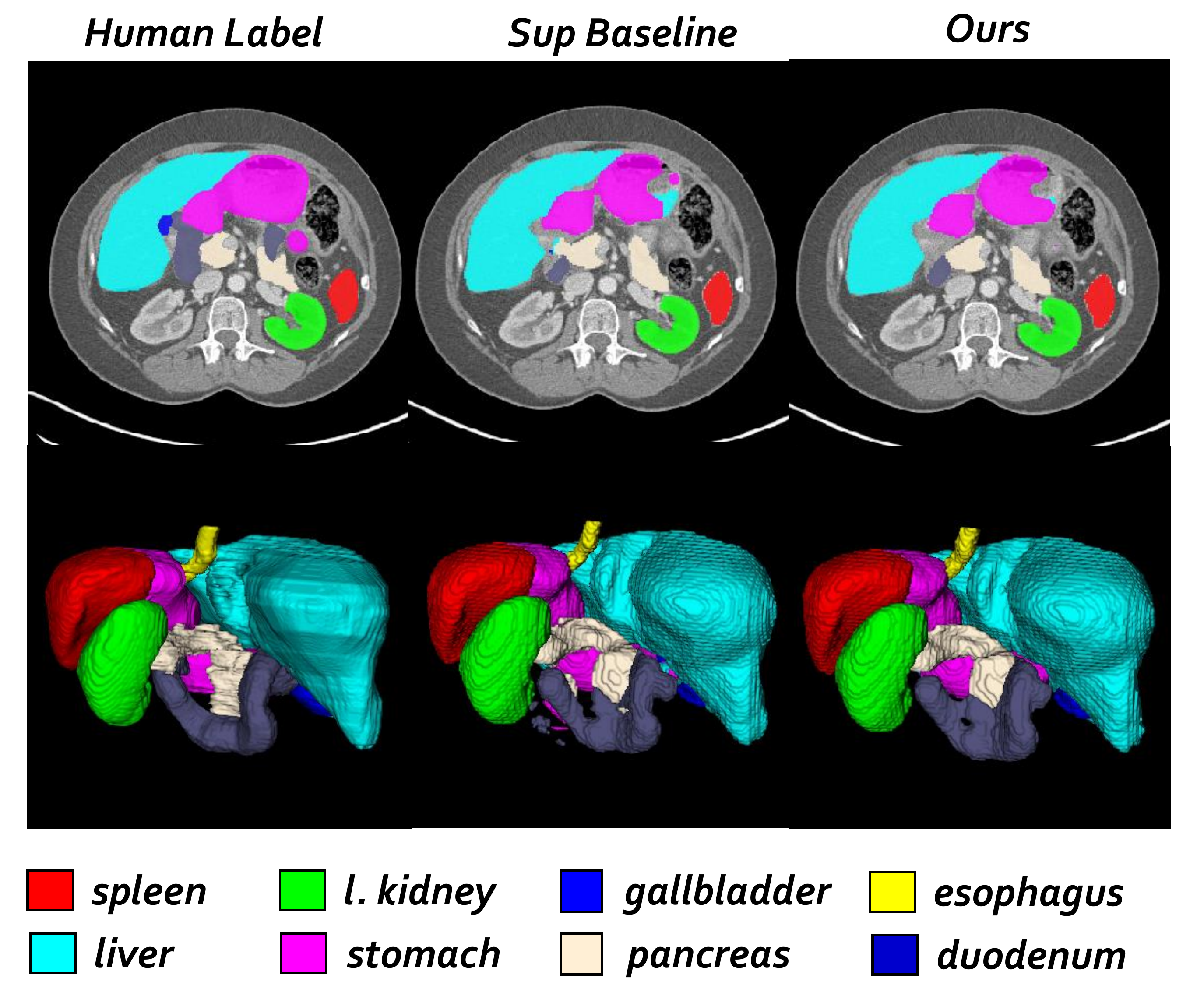}
\end{center}
\caption{
    An example of semi-supervised multi-organ segmentation.
}
\label{Fig:mo}
\end{figure}

\begin{figure*}[h]
\begin{center}
    \includegraphics[width=16cm]{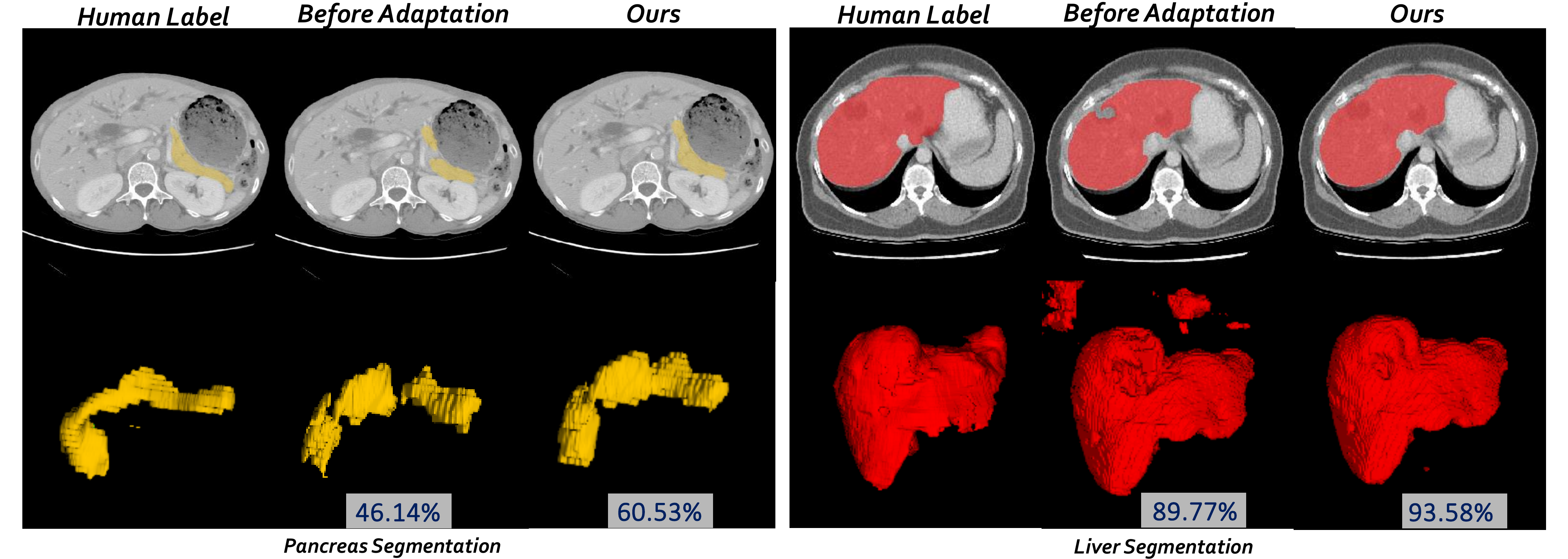}
\end{center}
\caption{
    2D and 3D visualizations for unsupervised domain adaptation of pancreas (left) and liver segmentation (right).
}
\label{Fig:uda}
\end{figure*}

\subsection{Multi-organ Segmentation Dataset}
Next, we validate our approach on multi-organ datasets. The dataset we use is a multi-organ re-annotated version from \cite{densevnet}, combining two public datasets - The Cancer Image Archive (TCIA) Pancreas-CT data set~\cite{roth2015deeporgan} and Beyond the Cranial Vault (BTCV) Abdomen data set~\cite{btcv}. \textcolor{revision}{We perform 4-fold cross validation on 90 cases in total. In each fold, we then randomly split the training cases into our labeled set $\mathcal{S}$ and unlabeled set $\mathcal{U}$. We train our models on different labeled data ratio 10\%, 20\%, which approximately corresponds to (7,81), (13,75) of (labeled, unlabeled) data pairs and validate on $\sim$22 cases in each fold.} 
Results are shown in Table~\ref{tab:multiorgan} and an example is shown in Fig~\ref{Fig:mo}.

Our approach improves consistently over almost every organ under every labeled-unlabeled ratio of data. The results illustrate the ability of our approach to handle the  situation of complex multi-organ settings.

\subsection{Unsupervised domain adaptation from multi-organ segmentation to MSD Dataset}
We aim to unsupervisedly adapt a model trained on TCIA multi-organ dataset to pancreas and liver cases in Medical Decathlon Challenge~\cite{msd} that can exhibit tumors. The target domains contain a shift from the source domain because of the differences in (i) image quality and contrast, and (ii) textures due to the existence of pancreatic/hepatic tumors. 

MSD pancreas dataset contains 282 CT scans in portal venous phase, all of which are pathological cases with pancreas and tumor annotation. We randomly split the whole dataset into 200 cases for training (without label) and 81 cases for validation. Since the source domain (multi-organ dataset) only contains healthy pancreas, we aim at segmenting the whole pancreas region (combining pancreas and tumor together). For MSD liver dataset, we aim at segmenting the whole liver region as well, with a random split of 100 training cases (unlabeled) and 31 cases for validation. 118 out of 131 cases contains hepatic tumor.

Table~\ref{Tab:uda} shows the results of unsupervised domain adaptation experiments. The first row is the segmentation performance (in terms of DSC) on pancreas and liver of the original multi-organ validation set. From the second row to the last, the results are DSC scores on MSD liver / pancreas validation set. In standard UDA settings (UMCT w/ source), significant improvements are achieved with our approach (1.12\% in liver and 4.70\% in pancreas), compared to source only version (direct Test on MSD). Due to the superior performance of self-training based approaches~\cite{cbst, crst} for unsupervised domain adaptation, we implement a vanilla self-training method under our settings (denoted as "Self-training"). \textcolor{revision}{We first test the model on the unlabeled set and then use the prediction as pseudo labels to train on the whole data set. We iterate these two steps every 1k iterations and trains for 5k iterations in total, which is in line with the proposed co-training scheme.} We also implement another baseline approach (AdaptSegNet~\cite{adaptsegnet}, denoted as "Adv training"), which applies adversarial training onto the predicted masks of semantic segmentation. \textcolor{revision}{The segmentation network serves as the generator to output segmentation masks with segmentation loss and tries to fool a patch-based discriminator (a 3D version of the discriminator used in AdaptSegNet~\cite{adaptsegnet}) with GAN loss~\cite{Goodfellow2015}. The discriminator is also trained jointly to distinguish between the predicted mask and the ground-truth mask on unlabeled data.} Our approach significantly outperforms all of the baselines. We also show one example for each organ in Fig~\ref{Fig:uda}.

The last row gives the results in the absence of source domain data. Under such condition, only a pre-trained source domain model and unlabeled target domain data are available.  Our UMCT-DA model (last row) is able to solve this problem by only using the co-training loss $L_{cot}$ to train on the target dataset, and achieve comparable results with standard UDA settings, even without source domain data. 

\begin{table}[!btp]
\centering
\begin{tabular}{|l|c|c|c|c|}    
\hline
Train & Test & Method & liver  & pancreas\\
\hline\hline
MO (L) & MO & Supervised &  95.59 & 81.69 \\
\hline\hline
MO (L)& MSD & Supervised & 92.78  & 70.23 \\
\hline\hline
\multirow{3}{*}{\thead{MO (L) \\+ \textcolor{revision}{MSD} (U)}} & \multirow{3}{*}{MSD} &  Adv training & 93.35 & 71.23\\
 & & Self-training & 92.67 & 71.38 \\
 & & UMCT &  \textbf{93.90} &  \textbf{74.93}\\

\hline\hline
\thead{MO model \\ + MSD (U)} & MSD& UMCT-DA& 92.98  & 74.38 \\
\hline
\end{tabular}
\caption{
    Experiments of unsuperivsed domain adaptation (UDA). The source domain is Multi-organ dataset (denoted as "MO") and target domains are MSD liver dataset and pancreas dataset. ``L" represents this dataset is labeled and ``U" means the opposite.
}
\label{Tab:uda}
\end{table}

\section{Discussions}
\subsection{Impact on large-scale benchmarks}
Under fully supervised training, our team NVDLMED was ranked the $3^{rd}$ place in the first phase and the \textbf{$\mathbf{2^{nd}}$ place in the final validation phase of Medical Segmentation Decathlon Challenge}~\cite{msd} (challenge leaderboard available\footnote{\url{http://medicaldecathlon.com/results.html}}). We applied our 3-view co-training framework taken from axial, coronal and sagittal views to ten medical image segmentation tasks simultaneously. The winning team~\cite{isensee2018nnu} applied heavy model selection and ensemble by cross-validation on the training set, while we used a fixed framework without complicated data augmentation. Although not originally targeted at improving the performance of fully supervised training, our approach still illustrated the effectiveness and robustness of co-training from multiple views. 

\subsection{Magnitude of domain shift}
In this work, domain shift mainly lies in various sources of CT scans and the pathological/healthy status of abdominal organs. \textcolor{revision}{We consider it a reasonable domain shift from different CT datasets originating from different hospitals and patient populations. Typically, this means that when directly transferring a model from one to another (TCIA to MSD in our case), the performance drops significantly (liver 95\% to 92\%, pancreas 81\% to 70\% average Dice). While this shift is relatively small compared to, for example, cross modality testing (say CT to MRI), it is unacceptable when considering these models for potential clinical applications. Considering the importance of this topic, we shed light on how well our semi-supervised approach performs on UDA tasks, given their similarity (discussed in Sec~\ref{Sec:prob}).}  Other types of domain shifts of medical images, though not investigated in this paper, are also of great importance. Investigation of domain adaptation under larger domain shifts such as modality changes (e.g. CT to MRI adaptation), contrast and resolution issues remains an active research topic.

\section{Summary \& Conclusion}
In this paper, we presented \textit{uncertainty-aware multi-view co-training} (UMCT), aimed at semi-supervised learning and domain adaptation. We extended dual view co-training and deep co-training into 3D volumetric image data by analysing from different view-points, then estimating uncertainty and finally enforcing multi-view consistency on large scale unlabeled data. Our approach was first validated on NIH pancreas dataset, where we outperformed other approaches by a large margin. We further applied our approach to multi-organ datasets and found significant improvements for each organ. Finally, we adapted the multi-organ dataset to MSD pathological pancreas and liver in an unsupervised manner. Our UMCT-DA model achieved good performance even in the absence of source domain data, illustrating strong potential for real-world applications in medical image segmentation. 

In the future, we plan to conduct further research in the following aspects. Currently the views of co-training are fixed and pre-defined, so one feasible idea is to incorporate more views and random views. This could increase the robustness of our model and lead to better performance. For domain adaptation, we will also try to explore co-training based approaches on other types of domain shifts including but not limited to image modality changes and contrast variants. We believe co-training based approaches will make a contribution to large scale medical image analysis with limited human annotations. 
\bibliography{refs}

\bibliographystyle{abbrv}

\end{document}